# Receipt Replay OOD: A Small Benchmark for Screen Replay Detection Under Domain Shift

*Alexander Vinogradov*
*IU International University of Applied Science*

*Dataset:*
*https://github.com/tehnix53/receipt-replay-ood*

**Abstract** - Public datasets such as DLC-2021, SynID, and KID34K have significantly contributed to research on presentation attack detection for identity documents, including screen replay attacks. However, evaluation of out-of-domain (OOD) robustness remains insufficiently explored, especially under realistic domain shifts.

In this work, we introduce *Receipt Replay OOD*, a small out-of-domain benchmark for screen replay detection. Receipts share several characteristics with identity documents, including planar geometry, curved corners, wear-and-tear artifacts, and text or logo patterns, while avoiding personally identifiable information constraints commonly associated with identity documents. We evaluate document replay detection models under cross-domain conditions and demonstrate the impact of domain shift on generalization performance. The dataset is publicly available.

## 1. Introduction and Related Works

Multiple datasets have been introduced for presentation attack detection in the domain of identity documents. The DLC-2021 dataset (Polevoy et al., 2022), based on images from the MIDV family, contains 10 document types and multiple attack categories including color photocopies, grayscale copies, and screen replay attacks captured under varying devices and lighting conditions. The SynID dataset (Tapia et al, 2025) approaches the problem from the perspective of structural realism and contains three document templates together with printed and screen replay attacks. The KID34K dataset (Eun-Ju et al., 2023) contains physical replicas of Korean identity documents designed to resemble real cards, together with corresponding screen and printed attacks.

Printed copy and screen replay attacks remain among the most important presentation attack types in identity document verification. However, transferability of printed copy detection approaches from academic datasets to real-world scenarios may have important limitations related to the complexity of real document manufacturing processes and the quality of bona fide samples available in public datasets. In practice, identity documents may use different substrates including secure paper and polycarbonate and employ a wide range of printing technologies such as offset, inkjet, dye-sublimation, and laser engraving, each introducing distinct visual and textural characteristics (Vinogradov, 2025). As a result, datasets focusing primarily on high-level design realism have inherent limitations for standalone training of robust production-grade printed copy detection systems. In contrast, screen replay attacks have a more clearly defined physical nature based on differences between direct captures of physical objects and images recaptured from electronic displays. Such recapturing processes introduce characteristic visual artifacts including moiré patterns, refresh-rate distortions, and frequency-domain inconsistencies. This makes screen replay attacks more suitable for studying model generalization and cross-domain robustness under controlled OOD conditions (Steinmann et al., 2024). A notable example is the work of Stehouwer et al. (2020), who introduced the GOSet dataset for screen recapture detection in a domain-agnostic setting containing bona fide and recaptured images of various generic objects such as keyboards, pens, and shoes.

## 2. Dataset description

Receipts were selected as lightweight real-world planar objects sharing several visual characteristics with identity documents, including text regions, curved geometry, folds, and wear-and-tear artifacts. At the same time, receipts avoid legal and privacy limitations associated with identity documents, making them suitable for public OOD benchmarking.

The benchmark contains both bona fide receipt captures and corresponding replay attack samples acquired under varied screen replay conditions. The dataset includes 497 bona fide images and 804 replay attack samples. Recaptured images were obtained using three mobile phones (Honor 8A, Samsung Galaxy J3, and Huawei P30 Lite) and two laptops (Lenovo ThinkBook 15 G2 and MacBook Pro M1). All samples are accompanied by concise metadata describing capture conditions such as indoor and outdoor environments and hand presence or absence as a possible source of bias. In addition, bounding box annotations are provided to localize receipt regions within each image.

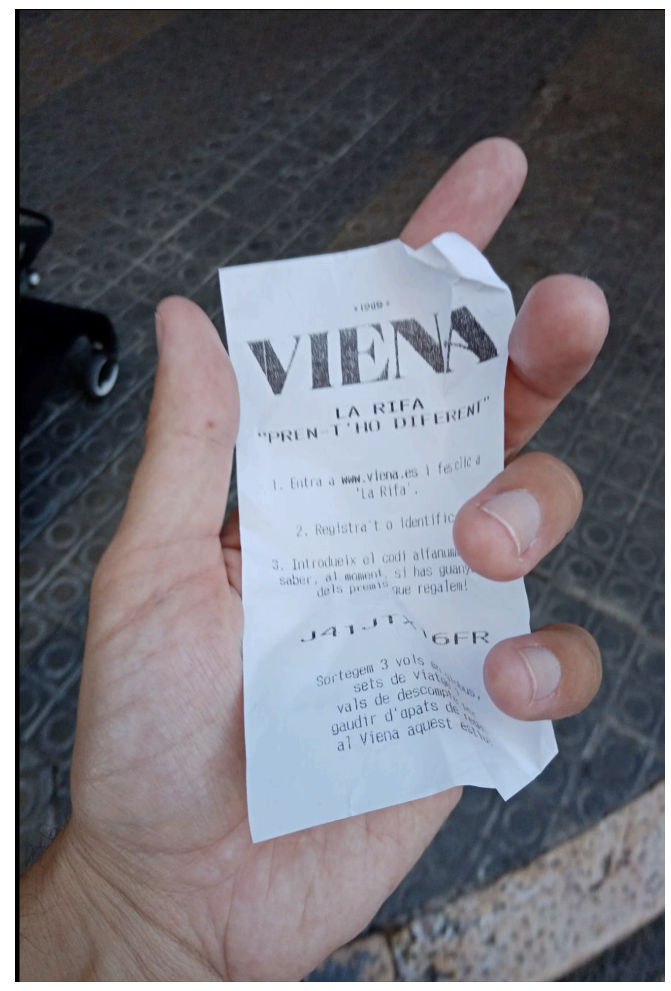

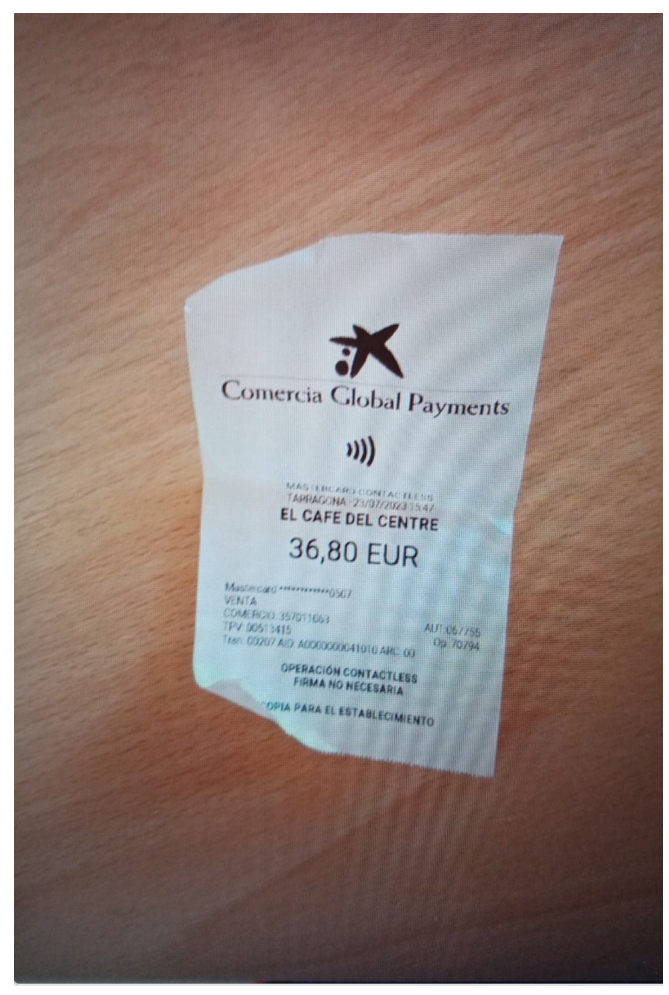

Fig. 1. Example bona fide (left) and screen replay attack (right) samples from the Receipt Replay OOD dataset.

## 3. Experiments

To support this research, we trained three models for the screen replay detection task using the original DLC-2021 (RE) train-test split. The evaluated architectures included a custom ResNet-inspired model trained from scratch, EfficientNet-B0V2 fine-tuned from ImageNet pretraining, and ViT-Small with frozen DINOv2 backbone and fine-tuned classification head.
Following the protocol proposed by the dataset authors, 224×224 patches were extracted from the central region of the document within the original frame. Training was performed using AdamW optimizer with cosine learning rate decay and binary cross-entropy loss. The DLC-2021 dataset exhibits class imbalance, with bona fide samples substantially outnumbering replay attack samples. To mitigate this issue, batch-level balancing was applied during training. Moderate augmentations including horizontal flips, small rotations, and brightness/contrast adjustments were applied. Models were trained or fine-tuned for up to 100 epochs with early stopping based on validation ROC AUC.
For each model, three independent runs were performed and the results were averaged. The obtained models were evaluated on both the original DLC-2021 (RE) testing partition and the proposed Receipt Replay OOD benchmark.

Table 1. ROC AUC degradation under domain shift

| | DLC ROC AUC | RR OOD ROC AUC | Δ ROC AUC |
|---|---|---|---|
| Custom CNN | 89.5 | 47.75 | -41.75 |
| EfficientNet-B0V2 | 88.45 | 65.32 | -23.14 |
| ViT-S/DINOv2 | 87.12 | 82.17 | -4.96 |

In the in-domain evaluation on the DLC-2021 test set, all three models demonstrated comparable performance, consistent with the results reported by the benchmark authors.
However, the out-of-domain evaluation on the proposed Receipt Replay OOD benchmark revealed substantial differences in robustness. The custom CNN trained from scratch exhibited severe performance degradation, suggesting reliance on domain-specific cues and shortcut correlations that failed to generalize beyond the training distribution. EfficientNet-B0V2 showed moderate sensitivity to domain shift, retaining partial discriminative ability despite noticeable degradation. In contrast, the DINOv2-based model demonstrated significantly stronger robustness under domain shift, achieving the best ROC AUC on the RR OOD dataset.

## 4. Conclusions and Future Work

OOD evaluation remains an important tool for assessing model robustness and generalization, especially in high-risk domains such as document verification. Among presentation attack scenarios, screen replay detection represents a particularly suitable task for OOD evaluation due to the relatively well-defined physical nature of replay artifacts introduced during screen replay.
In this work, we introduced Receipt Replay OOD, a small benchmark dataset for evaluating screen replay detection models under controlled domain shift conditions. Due to the planar nature of receipts and their visual similarity to document-like objects, the proposed benchmark may be particularly suitable for document-related replay detection tasks while remaining applicable to more general replay detection research. The proposed benchmark is based on the assumption that sufficiently generalized replay detection models should capture replay-specific visual artifacts that transfer across different categories of planar objects, including identity documents and receipts. Experimental results demonstrated substantial differences in cross-domain robustness between evaluated architectures and training paradigms. Although all models achieved comparable performance on the in-domain DLC-2021 benchmark, their performance degraded significantly on the proposed OOD dataset. These results further highlight the importance of alternative testing datasets for robustness evaluation beyond standard in-domain benchmarks. Unlike document imitation samples commonly found in academic datasets, the receipts in the proposed benchmark were collected under unconstrained everyday conditions rather than artificially prepared in laboratory settings. As a result, the data naturally contains folds, crumpling, tears, curved surfaces, and other imperfections that introduce additional variability and increase the complexity of replay detection.
The current benchmark has several limitations, including relatively limited dataset size and diversity, since it mainly contains black-and-white receipts collected from grocery stores, cafés, and markets. Future work may extend this approach toward broader categories of non-PII paper objects such as tickets, invoices, business cards, and other printed materials that remain visually closer to identity documents than generic object datasets.